\documentclass[sigconf,nonacm]{acmart}

\usepackage{graphicx}
\usepackage{multirow}
\usepackage{xcolor}
\usepackage{array}
\usepackage{tabularx}
\usepackage{adjustbox}
\usepackage{longtable}
\usepackage{subcaption}
\usepackage[most,breakable]{tcolorbox}

\definecolor{phcolor}{RGB}{180,0,0}
\definecolor{lightgray}{RGB}{240,240,240}
\definecolor{darkblue}{RGB}{31,73,125}
\definecolor{medblue}{RGB}{68,114,196}
\definecolor{lightblue}{RGB}{189,215,238}
\definecolor{darkgreen}{RGB}{55,96,35}
\definecolor{accentorange}{RGB}{197,90,17}

\newcommand{\sys}{ThuRunel}
\providecommand{\varnothing}{\emptyset}

\begin{document}

\title{ThuRunel: Dynamic Decoupling for Structured Advisory Dialogue}

\author{Yuyan Chen}
\authornote{Yuyan Chen is the Founder and CEO of \href{https://modelslive.org/}{ModelsLive Inc.}}
\affiliation{%
  \institution{ModelsLive Inc.}
  \country{}}
\email{ychen@modelslive.org}

\begin{abstract}
High-stakes advisory domains such as medical aesthetics, legal consultation, and educational planning exhibit a two-phase structure. The early phase requires empathetic elicitation and emotional support, and the late phase requires authoritative specialist judgment. Neither fully automated agents nor human junior consultants adequately address this structure at scale. We formalize the core design challenge as \emph{dynamic decoupling}, asking how an AI advisory agent should decide what to ask, when to stop, what to resolve autonomously, and what to forward to the specialist. We present \sys, an advisory agent combining a finite-state belief management framework, a chain-of-thought teacher synthesis protocol, and learned generation adapters. Against eleven baselines, \sys\ achieves consistent improvements in elicitation completeness and specialist brief quality. \sys\ is publicly deployed as a bilingual web application in which the same decoupling decisions operate from the client's side, grounded in a curated knowledge base that cites its sources in every answer.
\end{abstract}

\maketitle

\section{Introduction}
\label{sec:intro}

Advisory interactions in domains such as medical aesthetics, legal consultation, and educational counseling follow a characteristic two-phase structure \citep{chintagunta2021medically, topol2019high, rajpurkar2022ai}. The early phase is time-intensive but expertise-light, as clients present with partially formed preferences, conflicting self-sourced information, and anxiety requiring reassurance before meaningful elicitation can proceed \citep{zhou2024sotopia, liu2021towards}. The late phase is expertise-intensive but informationally compact, as the specialist needs a precise, distilled representation of the client's needs to render sound judgment. When senior specialists handle both phases, expert time is consumed by tasks requiring no specialist knowledge, and their empathy and availability are mismatched to anxious early-phase clients active outside business hours. Organizations address this with junior consultants, but this solution is costly, inconsistent, and difficult to scale \citep{nushi2018human, amershi2019software}.

\begin{figure}[t]
\centering
\includegraphics[width=0.9\linewidth]{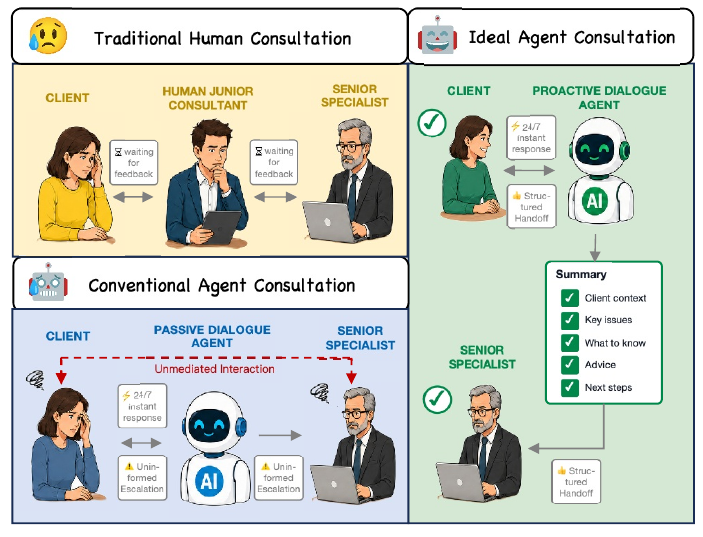}
\caption{Three advisory paradigms: Traditional human consultation is 
slow and office-hours-bound. Conventional agents escalate uninformed. The ideal 
advisory agent conducts structured elicitation and delivers a targeted handoff 
brief.}
\label{fig:overview}
\end{figure}

Conversational agents offer an alternative, but existing designs fail in one of two directions. Fully automated agents \citep{yao2023react, wu2023autogen} eliminate specialists from consequential decisions, while human-in-the-loop systems provide no throughput benefit \citep{mosqueira2023human, wu2022ai}. Task-oriented systems \citep{budzianowski2019hello, wu2019transferable, heck2020trippy} and prompted language models \citep{gpt55report, qwen3report} each address part of the problem but share a fundamental gap, as none models the partition between what the agent should resolve autonomously and what requires specialist judgment. As shown in Figure~\ref{fig:overview}, an ideal advisory agent conducts structured early-phase elicitation and delivers a targeted handoff brief, enabling the specialist to focus on consequential judgment.

We propose \emph{dynamic decoupling}, whereby the agent decides what to ask, when to stop eliciting, what to resolve within the dialogue, and what to forward to the specialist. When a client says ``my skin is sometimes sensitive,'' an existing agent accepts this and proceeds, leaving a critical contraindication unresolved. An agent following dynamic decoupling instead issues a targeted follow-up, asking ``Has your skin ever reacted to an injectable, such as significant swelling or prolonged redness?'' The client confirms a prior hyaluronic acid reaction. The agent resolves recovery and pricing questions within the dialogue and forwards to the specialist only the contraindication risk flag and matching service options.

We instantiate this paradigm in \sys\footnote{Live system: \url{https://thurunel.modelslive.org}.}, evaluated in the Chinese medical aesthetics market, where the industry reached RMB~312 billion in revenue in 2024 and continues to expand rapidly \citep{iresearch2024aesthetics}, making high-throughput advisory infrastructure a pressing operational need. \sys\ makes three contributions. First, we formalize dynamic decoupling as a multi-objective dialogue task with four agent decisions as the core challenge, and construct a training corpus through a chain-of-thought teacher synthesis protocol \citep{wei2022chain, ho2022large, fu2023specializing} that generates three-way parallel supervision from single base dialogues, enabling simultaneous training of elicitation, profile synthesis, and brief generation modules. Second, we propose an architecture combining a finite-state belief management framework \citep{young2013pomdp}, domain-adapted generation adapters, and programmatic specialist routing with deterministic boundary enforcement. Third, experiments against eleven baseline systems demonstrate that \sys\ has high specialist brief quality.

\section{Related Work}
\label{sec:related}

\paragraph{Task-oriented dialogue.}

\citet{budzianowski2019hello} establish belief-state tracking over typed
schemas as the foundational TOD paradigm, in which the dialogue state is a
flat slot-value mapping updated incrementally as the conversation
progresses. \citet{wu2019transferable} generate slot values via copy mechanisms that transfer across domains without schema-specific parameters, \citet{heck2020trippy} combine slot value copying, referencing, and confirmation in a single model, \citet{hosseini2020simple} show that a simple language model trained on task-oriented dialogues achieves competitive DST, and \citet{rastogi2020towards} propose the Schema-Guided Dialogue benchmark for zero-shot generalization to unseen services. \citet{xu2025schema} demonstrate that LLM-driven synthetic data
generation with step-by-step distillation achieves strong zero-shot DST,
and \citet{zhu2026multiturn} further show that multi-turn TOD synthesis
grounded in realistic task scenarios improves LLM reasoning in complex
dialogue settings, while \citet{mi2021cins} demonstrate that comprehensive instructions improve few-shot slot filling under noisy intent conditions. These approaches treat all slots as interchangeable and
leave phase transitions implicit in the generation policy, so they do not
address settings with tiered slot priorities, programmatic enforcement of
schema compliance and escalation actions, or a single dialogue that must
simultaneously support elicitation, profile synthesis, and specialist-facing
brief generation as coordinated objectives. \sys\ addresses all three gaps
through tier-stratified belief management, a programmatic pending-slot
register and action guard, and a teacher synthesis protocol that generates
aligned supervision for each objective from a single base dialogue.

\paragraph{Agent escalation.}
\citet{yao2023react} establish the foundational pattern of interleaved
reasoning and tool use for agents, and \citet{schick2024toolformer} show that language models can learn to invoke tools autonomously, but such agents execute tasks to
completion without a mechanism for handing off to a human when a decision
exceeds their scope; \citet{mosqueira2023human} survey human-in-the-loop machine learning and identify mixed-initiative escalation as the key design axis for such settings. \citet{wu2024collabllm} demonstrate that LLMs trained
with next-turn rewards remain passive responders in multi-turn settings,
motivating our use of chain-of-thought teacher synthesis, where \citet{wei2022chain} show that step-by-step reasoning elicitation substantially improves multi-step task performance and \citet{wang2023self} further show that aggregating multiple reasoning paths improves reliability, over standard SFT.
\citet{feng2024rehac} show that reinforcement learning for optimal human
intervention timing outperforms both full automation and turn-by-turn
oversight, and \citet{ouyang2022training} demonstrate that training with human feedback aligns agent behavior to intended boundaries; \citet{wang2026ahce} further show that effective human-AI
collaboration requires learning when and how to request expert input, and \citet{wu2022ai} establish that explicit boundary communication between AI and human decision-makers is a prerequisite for trust calibration. For
evaluation, \citet{zheng2023judging} develop the LLM-as-Judge framework
that we also adapt to the structured advisory setting.

\begin{figure*}[t]
\centering
\includegraphics[width=0.9\textwidth]{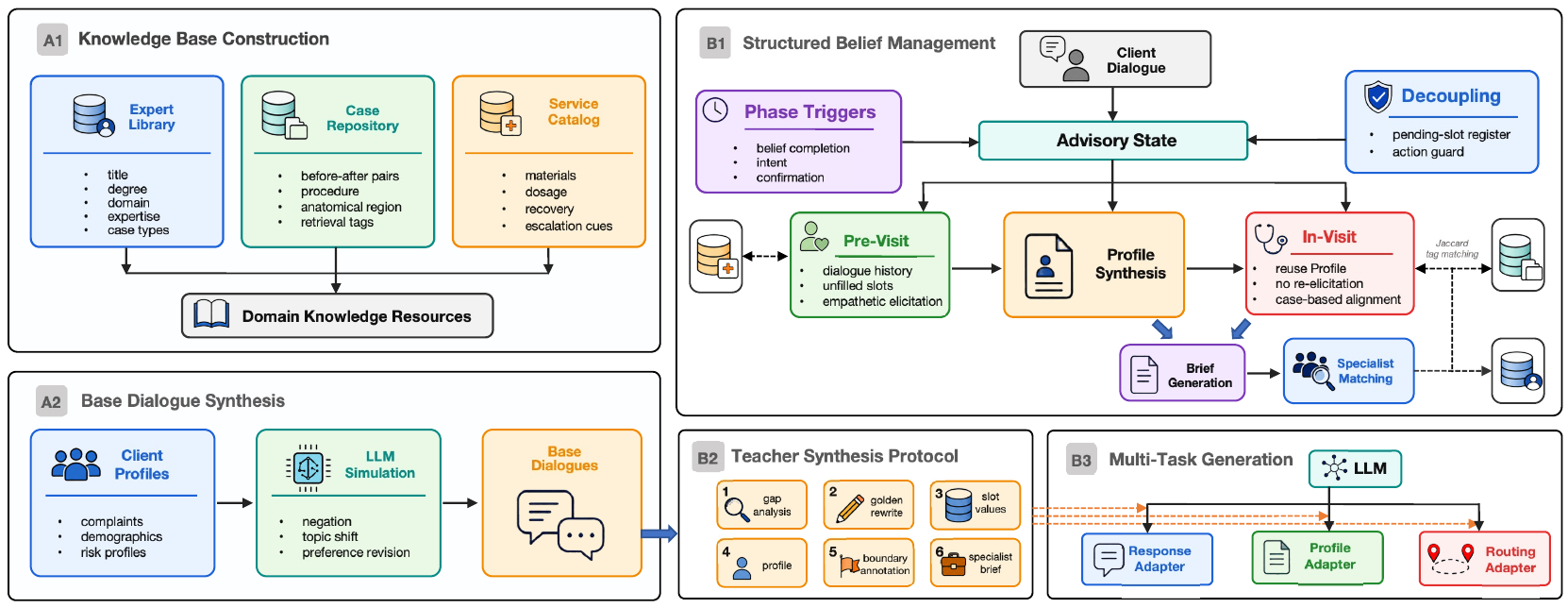}
\caption{Dataset construction and \sys\ framework architecture. A1--A2 cover
knowledge resource compilation and base dialogue synthesis. B1--B3 show the
framework components: structured belief management, the six-step teacher synthesis protocol, and the three LLM adapters trained on the resulting supervision.}
\label{fig:framework}
\end{figure*}

\section{Task Formulation}
\label{sec:problem}

Let $\mathcal{D} = \{(u_1, a_1), \ldots, (u_T, a_T)\}$ be a multi-turn advisory dialogue where $u_t$ is a client utterance and $a_t$ is the agent response. The agent maintains a belief state $b_t : \mathcal{S} \to \mathcal{V} \cup \{\varnothing\}$ updated by extraction function $\phi$:
\begin{equation}
b_{t+1}(s) = \begin{cases} \phi(u_t, s) & \phi(u_t, s) \neq \varnothing \\ b_t(s) & \text{otherwise} \end{cases}
\label{eq:belief}
\end{equation}
The paradigm decomposes the interaction into four agent decisions. $\mathcal{T}_a$ is turn generation, producing $a_t$ from dialogue prefix $\mathcal{D}_{<t}$ and belief state $b_t$ under tier-priority ordering while sustaining empathetic register. $\mathcal{T}_s$ is phase transition, firing the handoff at $T^* = \min\{t : b_t(s) \neq \varnothing \text{ for all } s \in \mathcal{S}_1\}$ where $\mathcal{S}_1$ is the Tier-1 slot set. $\mathcal{T}_p$ is profile synthesis, generating client profile $P$ from $\mathcal{D}$ and $b_{T^*}$ as the handoff record. $\mathcal{T}_b$ is brief generation, producing specialist brief $B$ from $P$, $\mathcal{D}$, and knowledge base $\mathcal{K}$, presenting only specialist-actionable content with advisory-scope information already resolved within the dialogue.
The slot schema $\mathcal{S}$ comprises 12 typed fields in three tiers defined in Table~\ref{tab:schema}. We report $\mathrm{SFR} = \frac{1}{|\mathcal{S}|}\sum_{s}\mathbf{1}[b_T(s)\neq\varnothing]$ and tier-stratified variants SFR$_1$, SFR$_2$, SFR$_3$.

\begin{table}[!htbp]
\centering
\small
\begin{tabularx}{\linewidth}{lXX}
\toprule
Tier & Definition & Example slots \\
\midrule
1 & Highest-priority, decision-critical information that should be completed before recommendation or specialist handoff. & Chief complaint, target anatomy, contraindications, budget \\
2 & Preference and feasibility constraints that refine the advisory process after Tier-1 information is covered. & Desired effect, recovery window, risk preference, procedure interest \\
3 & Supplementary and logistical information used for profile completion, appointment arrangement, and follow-up. & Skin type, medical history, appointment window, contact information \\
\bottomrule
\end{tabularx}
\caption{Slot schema grouped by tier-priority assignments. The schema, state-machine transitions, and tier-priority ordering are all specified through configuration files, enabling adaptation to other advisory domains such as education advising or legal triage by substituting domain-specific slots and boundary triggers.}
\label{tab:schema}
\end{table}

\section{Dataset Construction}
\label{sec:data}
The dataset construction process is shown in Figure~\ref{fig:framework} (A1--A2).

\paragraph{Knowledge base construction.}
Three types of knowledge resources are compiled prior to dialogue data generation. The expert library $\mathcal{K}_E$ contains structured specialist profiles with fields for clinical title, degree, primary domain, procedure expertise, and representative case types. Profiles are authored by clinic administrators and validated by a senior practitioner. The case repository $\mathcal{K}_C$ is indexed from a collection of annotated before-and-after image pairs, each record linking image paths to procedure category, anatomical region, and a tag set used for retrieval. The service catalog $\mathcal{K}_S$ encodes procedure-specific advisory knowledge, including product material recommendations, dosage guidance by anatomical sub-region, recovery expectation templates, and escalation cues for contraindication-sensitive scenarios, compiled from clinical guidelines and practitioner protocols.

\paragraph{Base dialogue synthesis.}
Domain practitioners specify 2000 representative client profiles spanning the distribution of chief complaints, demographics, and risk profiles in medical aesthetics practice. For each profile, an initial multi-turn dialogue is simulated by prompting a language model with a detailed persona-instantiating instruction that elicits realistic early-phase conversational behavior, including colloquial negation, topic redirection, and preference revision. These base dialogues serve as inputs to the teacher synthesis protocol and are not used as training targets.

\paragraph{Dataset statistics.} The final training corpus contains 2000 pre-visit dialogue sessions and 2000 specialist-brief sessions, generated from the
2000 client profiles described above. These data are used to train the
client-facing response adapter, the handoff-generation adapter, and the
routing adapter for task selection. An additional 400 sessions form the
held-out evaluation set.

\begin{table*}[t]
\centering
\small
\resizebox{0.7\linewidth}{!}{%
\begin{tabular}{llcccccccccc}
\toprule
& & \multicolumn{4}{c}{LLM Judge Score} & \multicolumn{6}{c}{Human $J_S$ by Dimension} \\
Grp & System & $J_T$ & $J_S$ & SFR & BIV & Emp & Comp & Acc & Struct & Bound & Avg. \\
\midrule
\multirow{5}{*}{A} & GPT-5.5 & 85.7 & 80.2 & 80.3 & \underline{84.6} & \underline{88.7} & 86.1 & 86.3 & 82.5 & \underline{85.3} & \underline{85.8} \\
 & Gemini-3.1 & 76.2 & 71.8 & 72.7 & 65.7 & 85.2 & 83.3 & 79.5 & 73.1 & 72.5 & 78.7 \\
 & Grok-4.20 & 81.3 & 72.4 & 75.2 & 72.5 & 78.5 & \underline{88.3} & \underline{88.3} & 78.9 & 78.1 & 82.4 \\
 & DeepSeek-V4 & 74.5 & 79.2 & 74.1 & 63.8 & 87.1 & 83.5 & 83.5 & 75.2 & 77.5 & 81.4 \\
 & Claude-4.6 & \underline{88.2} & \underline{82.5} & \underline{82.4} & 80.2 & 72.5 & \underline{91.1} & 84.6 & \underline{84.5} & 79.4 & 82.4 \\
\midrule
\multirow{4}{*}{B} & Qwen3-32B & 72.6 & 73.3 & 70.2 & 54.7 & 71.4 & 77.6 & 77.7 & 75.6 & 76.6 & 75.8 \\
 & GLM-4.5 & 73.3 & 70.8 & 68.3 & 57.1 & 68.6 & 70.3 & 73.5 & 73.3 & 70.2 & 71.2 \\
 & Yi-1.5-34B & 70.5 & 67.3 & 66.2 & 60.3 & 70.4 & 68.4 & 66.1 & 66.5 & 66.7 & 67.6 \\
 & InternLM-2.5 & 68.8 & 70.9 & 63.1 & 54.8 & 72.1 & 63.3 & 69.3 & 68.9 & 69.5 & 68.6 \\
\midrule
\multirow{2}{*}{C} & ReAct & 73.4 & 72.1 & 65.4 & 66.5 & 64.5 & 64.9 & 57.4 & 63.1 & 71.4 & 64.3 \\
 & AutoGen & 78.1 & 75.5 & 66.2 & 72.8 & 67.7 & 66.5 & 58.6 & 65.7 & 72.7 & 66.2 \\
\midrule
\multirow{1}{*}{D} & \sys & \textbf{92.7} & \textbf{90.2} & \textbf{88.3} & \textbf{91.4} & \textbf{94.3} & \textbf{95.2} & \textbf{93.5} & \textbf{95.3} & \textbf{96.0} & \textbf{94.9} \\
\bottomrule
\end{tabular}}
\caption{Main results from the LLM judge and human evaluation. 
The metrics $J_T$ and $J_S$ are turn level and session level LLM judge scores. SFR is slot fill rate. BIV is brief information validity. Emp is empathy. Comp is completeness. Acc is accuracy. Struct is structure. Bound is boundary compliance. Avg. is the average of the five human dimensions. Bold indicates the best result in each column and underline indicates the second-best.}
\label{tab:main}
\end{table*}

\begin{table}[t]
\centering
\small
\resizebox{0.86\linewidth}{!}{%
\begin{tabular}{llcccc}
\toprule
Grp & System & SFR$_1$ & SFR$_2$ & SFR$_3$ & Eff \\
\midrule
A & GPT-5.5 & 78.2 & 73.8 & 88.9 & 8.5 \\
B & Qwen3-32B & 68.9 & 58.1 & 83.6 & 13.4 \\
C & AutoGen & 63.3 & 57.8 & 77.5 & 15.7 \\
D & \sys & 90.3 & 82.2 & 92.4 & 5.2 \\
\bottomrule
\end{tabular}}
\caption{Tier stratified slot fill rate and dialogue efficiency. The metrics SFR$_1$, SFR$_2$, and SFR$_3$ measure fill rates for Tier 1, Tier 2, and Tier 3 slots. Eff reports the average number of turns required to complete Tier 1 slots.}
\label{tab:diagnostic}
\end{table}

\begin{table*}[t]
\centering

{\scriptsize

\setlength{\tabcolsep}{3pt}
\renewcommand{\arraystretch}{1.0}
\setlength{\aboverulesep}{0.8pt}
\setlength{\belowrulesep}{0.8pt}

\begin{adjustbox}{max width=\textwidth,max totalheight=\textheight,keepaspectratio}
\begin{tabular}{p{0.16\textwidth}p{0.78\textwidth}}
\toprule
Metric & Scoring standard \\
\midrule
$J_T$, $J_S$, Emp, Comp, Acc, Struct, Bound &
90--100: Excellent. Fully satisfies the criterion with no major issues.\newline
80--89: Good. Satisfies the criterion with only minor issues.\newline
70--79: Adequate. Mostly usable but has noticeable weaknesses.\newline
60--69: Weak. Partially satisfies the criterion but has substantial omissions or quality problems.\newline
0--59: Poor. Fails the criterion or contains severe errors. \\
\midrule
SFR, SFR$_1$, SFR$_2$, SFR$_3$ &
90--100: Excellent. Nearly all required slots in this category are correctly filled.\newline
80--89: Good. Most required slots in this category are correctly filled, with minor omissions.\newline
70--79: Adequate. A majority of required slots are filled, but several important fields are missing.\newline
60--69: Weak. Only partial slot recovery is achieved, with substantial missing information.\newline
0--59: Poor. Slot recovery is largely incomplete or unreliable. \\
\midrule
Eff &
Lower is better. The value reports the average number of turns required to complete all Tier-1 slots.\newline
A smaller value indicates more efficient elicitation of high-priority information.\newline
A larger value indicates slower elicitation or unnecessary dialogue turns before Tier-1 completion. \\
\midrule
BIV &
90--100: Excellent. Nearly all brief content is actionable for the specialist.\newline
80--89: Good. Most brief content is actionable, with minor non-actionable details.\newline
70--79: Adequate. The brief contains useful information but also noticeable irrelevant or advisory-scope content.\newline
60--69: Weak. A substantial portion of the brief is non-actionable, redundant, or poorly targeted.\newline
0--59: Poor. The brief is largely non-actionable or fails to support specialist decision making. \\
\midrule
Cls &
90--100: Excellent. The system clearly identifies the difficult response, asks a targeted follow-up, and fully recovers the intended slot value.\newline
80--89: Good. The system asks an appropriate follow-up and recovers the intended slot value with only minor ambiguity.\newline
70--79: Adequate. The system partially clarifies the response and recovers enough information for a usable belief-state update.\newline
60--69: Weak. The system attempts clarification, but the recovered information remains incomplete or uncertain.\newline
0--59: Poor. The system fails to clarify the difficult response or does not recover the intended slot value. \\
\midrule
Rep &
0--5: Excellent. No unnecessary repeated questions or only negligible redundancy.\newline
6--10: Good. Very limited repetition that does not meaningfully affect the dialogue.\newline
11--20: Adequate. Some repeated questioning occurs, but the dialogue remains mostly usable.\newline
21--40: Weak. Repeated questioning is noticeable and affects user experience or dialogue efficiency.\newline
41--100: Poor. Severe repetition, with the system repeatedly asking for information already elicited or credibly answered. \\
\bottomrule
\end{tabular}
\end{adjustbox}
}

\caption{LLM evaluation scoring standards. Metrics with identical scoring criteria are grouped in the same row.}
\label{tab:llm_scoring}

\end{table*}

\begin{table}[t]
\centering

{\fontsize{7.5pt}{10pt}\selectfont
\setlength{\tabcolsep}{4pt}
\renewcommand{\arraystretch}{0.95}
\setlength{\aboverulesep}{0.8pt}
\setlength{\belowrulesep}{0.8pt}

\begin{tabularx}{\columnwidth}{X}
\toprule
Scoring standard \\
\midrule
90--100: Excellent. The experience strongly satisfies the criterion.\newline
80--89: Good. The experience satisfies the criterion with minor issues.\newline
70--79: Adequate. The experience is acceptable but has weaknesses.\newline
60--69: Weak. The experience only partially satisfies the criterion.\newline
0--59: Poor. The experience fails the criterion or is clearly unsatisfactory. \\
\bottomrule
\end{tabularx}
}

\caption{Human evaluation scoring standards. The same scale is used for Nat, Und, Trust, Will, and Overall.}
\label{tab:human_scoring}

\end{table}

\section{The \sys\ Framework}
\label{sec:system}

The \sys\ framework is shown in Figure~\ref{fig:framework} (B1--B3).

\subsection{Structured Belief Management}

\sys\ represents the advisory state as a triple $(q, b_t, \tau)$ where $q \in Q$ is the current phase, $b_t$ is the belief state from Equation~\ref{eq:belief}, and $\tau$ is a turn counter. Phase transitions follow a deterministic function $\delta: Q \times \Sigma \to Q$ under three trigger classes, of which belief-completion triggers fire when all slots in $\mathcal{S}_q$ are filled, intent triggers fire on predefined semantic patterns such as in-clinic arrival, and confirmation triggers respond to system-level events such as booking persistence. Two mechanisms enforce the decoupling decisions. The pending-slot register records elicitation attempts in turn $t$ and applies a normalization mapping at turn $t{+}1$ that resolves colloquial null expressions to the queried slot, encoding $\mathcal{T}_s$ as a post-hoc belief update. The action guard intercepts outputs matching booking confirmation or diagnostic assertion patterns and substitutes a programmatically issued token, preventing hallucinated commitments.

The pre-visit phase realizes $\mathcal{T}_a$ via prompt-conditioned generation: $\theta_u$ receives the dialogue history, tier-stratified unfilled slots, and a phase-indicator token, prioritizing Tier-1 elicitation through in-context schema enforcement. Phase completion triggers profile generation under $\mathcal{T}_p$. The in-visit phase inherits $P$, suppresses re-elicitation of filled slots, and implements case-based expectation alignment as RAG over $\mathcal{K}_C$ using Jaccard similarity $|T_q \cap T_c|/|T_q \cup T_c|$. The brief module $\theta_p$ generates $B$ under $\mathcal{T}_b$, and tag-overlap matching routes to the appropriate specialist in $\mathcal{K}_E$.

\subsection{Teacher Synthesis Protocol}
\label{sec:cot}

GPT-5.2, the strongest model available at the time of dataset construction, executes a six-step chain-of-thought protocol per base dialogue, transforming it into three parallel supervision signals. Step~1 performs gap analysis over unfilled slots, premature recommendations, and register violations. Step~2 produces a golden rewrite, whose assistant turns supervise $\theta_u$. Step~3 enumerates slot values with provenance and Step~4 compiles the typed client profile $P$, supervising $\theta_p$ on profile synthesis. Step~5 annotates the $\mathcal{T}_b$ boundary between advisory-scope and specialist-scope content, and Step~6 generates brief $\widehat{B}$ conditioned on that annotation, supervising $\theta_p$ on brief generation. A single execution thus yields supervision for all three adapter objectives from one base dialogue, without separate annotation campaigns. Three medical graduate student volunteers manually reviewed a random sample of the synthesized supervision, rating each on a 1--5 scale for correctness and usability, with all sampled instances scoring above 4.

\subsection{Multi-Task Generation}

Three adapters are learned via QLoRA \citep{dettmers2023qlora} over a shared Qwen2.5-32B-Instruct backbone using next-token prediction on assistant-role tokens. $\theta_u$ learns empathetic elicitation from turn-level rewrites. $\theta_p$ learns the advisory-to-specialist content boundary from brief supervision. $\theta_r$ learns task-type classification using prefix tokens $\tau_a$, $\tau_p$, $\tau_b$, with weight updates parametrized as $W = W_0 + \frac{\alpha}{r}BA$ \citep{hu2022lora}. Adapter selection at inference time is handled deterministically by the belief management layer.

\begin{table}[t]
\centering
\small
\resizebox{0.7\linewidth}{!}{%
\begin{tabular}{lcccc}
\toprule
Configuration & $J_T$ & $J_S$ & SFR & BIV \\
\midrule
\sys & 92.7 & 90.2 & 88.3 & 91.4 \\
w/o CoT & 84.3 & 81.8 & 80.1 & 82.1 \\
w/o Rou & 88.9 & 85.2 & 85.3 & 85.3 \\
w/o PSR & 82.2 & 80.5 & 75.3 & 81.2 \\
w/o BM    & 79.1 & 77.2 & 74.1 & 80.9 \\
\bottomrule
\end{tabular}}
\caption{Ablation results. w/o CoT removes CoT augmentation. w/o Rou removes the task routing adapter. w/o PSR removes the pending slot register. w/o BM removes the full belief management framework, degrading to a prompt-only baseline.}
\label{tab:ablation}
\end{table}

\begin{table}[t]
\centering
\small
\resizebox{0.9\linewidth}{!}{%
\begin{tabular}{llccccc}
\toprule
Grp & System & Nat & Und & Trust & Will & Overall \\
\midrule
A & GPT-5.5 & 4.4 & 3.7 & 4.1 & 3.8 & 4.2 \\
B & Qwen3-32B & 3.8 & 3.3 & 3.8 & 3.5 & 3.5 \\
C & AutoGen & 3.2 & 3.1 & 3.9 & 3.1 & 3.2 \\
D & \sys & 4.5 & 4.4 & 4.5 & 4.3 & 4.5 \\
\bottomrule
\end{tabular}}
\caption{Client satisfaction from role playing evaluators. Nat is naturalness. Und is sense of being understood. Will is willingness to complete.}
\label{tab:satisfaction}
\end{table}

\begin{table}[t]
\centering
\small
\setlength{\tabcolsep}{4pt}
\resizebox{0.85\linewidth}{!}{%
\begin{tabular}{llcccc}
\toprule
Cond. & Sys & $J_T$ & SFR$_1$ & Cls & Rep \\
\midrule
\multirow{2}{*}{Direct} & GPT-5.5 & 95.2 & 88.6 & - & 3.2 \\
 & \sys & 97.1 & 96.7 & - & 1.3 \\
\midrule
\multirow{2}{*}{Underspecified} & GPT-5.5 & 90.4 & 81.7 & 64.8 & 18.5 \\
 & \sys & 93.5 & 93.8 & 79.2 & 9.3 \\
\midrule
\multirow{2}{*}{Deflective} & GPT-5.5 & 82.3 & 70.4 & 55.7 & 24.2 \\
 & \sys & 90.3 & 89.6 & 73.8 & 11.4 \\
\midrule
\multirow{2}{*}{Inconsistent} & GPT-5.5 & 74.9 & 72.1 & 42.1 & 16.3 \\
 & \sys & 89.9 & 81.1 & 70.5 & 5.5 \\
\bottomrule
\end{tabular}
}
\caption{Robustness under controlled user response conditions. Direct serves as the reference condition. $J_T$ is the turn level judge score. SFR$_1$ measures recovery of high priority slots. Cls measures clarification success and is not applicable to Direct responses. Rep measures repeated question rate.}
\label{tab:robustness}
\end{table}

\section{Experiments}
\label{sec:exp}

\subsection{Experimental Setup}

All adapters are trained on a single 80\,GB A100 GPU with learning rate $2 \times 10^{-4}$ and effective batch size 4. We compare against eleven systems in four groups. Group A comprises five frontier closed-source models prompted with the full domain prompt. Group B comprises four open-source models fine-tuned on raw advisory dialogues without CoT synthesis, matching \sys's base capacity but without the teacher protocol. Group C comprises a ReAct-style system over GPT-5.5 with slot-fill, booking, and retrieval tools, and an AutoGen two-agent system.
We evaluate on the held-out set of 400 sessions. Automatic metrics $J_T$, $J_S$, SFR, and BIV are scored by Kimi~2.6 as judge. Human evaluation dimensions in Table~\ref{tab:main} are rated by three annotators with inter-annotator agreement Krippendorff's $\alpha$ at 0.76. Client satisfaction in Table~\ref{tab:satisfaction} is collected from five role-playing evaluators per session on a 1--5 Likert scale. These evaluators are graduate students without medical training to simulate the perspective of a typical client.

\paragraph{Metric definitions.}
$J_T$ (Turn-level Judge Score) measures the overall quality of each system response at the turn level, evaluating helpfulness, contextual appropriateness, naturalness, informativeness, and alignment with the advisory goal. $J_S$ (Session-level Judge Score) measures overall quality of the full advisory session, evaluating whether the system completes the consultation effectively, maintains coherent dialogue flow, collects necessary information, respects professional boundaries, and produces a useful outcome. SFR (Slot Fill Rate) measures the percentage of required slots successfully filled by the end of the dialogue. BIV (Brief Information Validity) measures the proportion of generated specialist-brief content that is actionable for the specialist, counting information that supports risk assessment, treatment feasibility, constraint verification, or specialist routing. Human evaluation dimensions measure Empathy (Emp), Completeness (Comp), Accuracy (Acc), Structure (Struct), and Boundary Compliance (Bound), each on a 0--100 scale. Client satisfaction dimensions measure Naturalness (Nat), Sense of Being Understood (Und), Trust, Willingness to Complete (Will), and Overall satisfaction. 
All judge-based metrics are scored on a 0--100 scale unless otherwise specified. LLM-based metrics are scored by Kimi 2.6 using task-specific prompts, while human-facing metrics are scored by volunteer annotators using the corresponding questions. Rep is a reverse-direction metric, where lower values indicate fewer repeated questions. Score-band definitions used for LLM and human evaluation are in Table~\ref{tab:llm_scoring} and Table~\ref{tab:human_scoring}.

\begin{figure}[!htbp]
\begin{tcolorbox}[colback=lightgray, colframe=darkblue!50, fontupper=\small]
\textbf{Client:} Hello. I would like to learn about methods for making my face look slimmer.

\medskip
\textbf{\sys:} Hello! I am a medical aesthetics intake assistant. May I ask your name?

\medskip
\textbf{GPT-5.5:} There are several ways people make the face appear slimmer, ranging from temporary styling techniques to longer-term cosmetic procedures. What works best depends on your facial structure, body composition, age, and how dramatic a change you want.
\end{tcolorbox}
\caption{Turn 1: opening. \sys\ begins structured elicitation. GPT-5.5 lists methods without collecting any typed slot.}
\label{fig:t1_body}
\end{figure}

\begin{figure}[!htbp]
\begin{tcolorbox}[colback=lightgray, colframe=darkblue!50, fontupper=\small]
\textbf{Client:} Budget around RMB 15,000. Skin-wise I am just sometimes a bit sensitive.

\medskip
\textbf{\sys:} You mentioned sensitive skin. Has your skin ever had a strong reaction to an injectable treatment, such as significant swelling or lasting redness afterward?

\medskip
\textbf{GPT-5.5:} With a budget of around RMB 15,000 and mild skin sensitivity, you have access to several good low-downtime options. The safest and most effective choice still depends on what is making your face appear wider.
\end{tcolorbox}
\caption{Turn 4: contraindication elicitation. \sys\ issues a targeted pending-slot follow-up. GPT-5.5 treats the mention as a preference and does not resolve the contraindication slot.}
\label{fig:t4_body}
\end{figure}

\begin{figure}[!htbp]
\begin{tcolorbox}[colback=lightgray, colframe=darkblue!50, fontupper=\small]
\textbf{Client:} Actually yes, I tried hyaluronic acid a few years ago and had a bad reaction.

\medskip
\textbf{\sys:} That is really important to know. A prior reaction to hyaluronic acid will shape which approaches are appropriate for you. I have noted this in your record and our specialist will want to discuss this carefully.

\medskip
\textbf{GPT-5.5:} That's important to know. A bad reaction to hyaluronic acid (HA) filler changes how you should approach facial treatments, especially injections.
\end{tcolorbox}
\caption{Turn 5: contraindication confirmed. \sys\ records the prior reaction and flags it for the specialist. GPT-5.5 immediately offers clinical guidance, crossing the advisory boundary.}
\label{fig:t5_body}
\end{figure}

\begin{figure}[!htbp]
\begin{tcolorbox}[colback=lightgray, colframe=darkblue!50, fontupper=\small]
\textbf{Client:} I would like to proceed with the balanced plan.

\medskip
\textbf{\sys\ (specialist brief):} Client age: 20. Chief concern: facial slimming, non-surgical preference. Contraindication: prior hyaluronic acid reaction. Selected plan: balanced plan, medium dosage. Recovery window: 5--7 days. Estimated cost: RMB 3000--4000. Routing: Dr.~[matched specialist], facial aesthetics and injectable focus.

\medskip
\textbf{GPT-5.5:} Great, for your goals, budget, short recovery preference, and previous HA reaction, I'd suggest a balanced, conservative plan rather than aggressive reshaping.
\end{tcolorbox}
\caption{Turn 9: specialist brief. \sys\ generates a structured handoff with contraindication flag and specialist routing implementing $\mathcal{T}_b$. GPT-5.5 produces a conversational suggestion without structured handoff or routing.}
\label{fig:t9_body}
\end{figure}

\subsection{Main Results}

Table~\ref{tab:main} shows that \sys\ achieves the strongest performance across all automatic and human metrics, exceeding the best baseline on every column. Among baselines, Claude-4.6 leads on $J_T$ and SFR while GPT-5.5 leads on BIV and human average. The gains confirm that \sys\ improves not only information recovery and brief quality but also turn-level response quality and human-perceived dialogue quality.
Table~\ref{tab:diagnostic} shows the gains are concentrated in Tier-1: \sys\ reaches SFR$_1$ at 90.3 versus the best baseline at 78.2, and completes Tier-1 in 5.2 turns versus 8.5, confirming more efficient elicitation of decision-critical information.

We compute Pearson correlations across the 400 evaluation sessions for \sys.
$J_T$ and $J_S$ are strongly correlated, with Pearson $r$ at 0.872, showing that turn level quality and session level quality are aligned. $J_S$ also correlates strongly with the average of the five human dimensions, with $r$ at 0.801. Among the individual human dimensions, $J_S$ correlates most strongly with boundary compliance, with $r$ at 0.921, followed by structure at 0.840, completeness at 0.737, empathy at 0.659, and accuracy at 0.638. This suggests that, in this advisory setting, boundary compliance and response structure are the human evaluated dimensions most closely associated with overall session quality.

\subsection{Ablation Study}

Table~\ref{tab:ablation} shows that removing CoT produces the largest quality drop, with $J_T$ falling from 92.7 to 84.3 and BIV from 91.4 to 82.1, confirming the
teacher protocol as the primary contributor. Removing PSR produces the
largest SFR drop (88.3 to 75.3), with an accompanying BIV decline as
incomplete slot coverage leaves the brief generator with a partially
populated belief state. Removing BM produces the largest combined drop,
confirming that PSR, action guard, and FSM transitions operate as an
integrated unit, while removing Rou produces the smallest drop across all
four metrics. The routing adapter $\theta_r$, evaluated on a held-out 10\%
split, achieves 94.9\% overall accuracy with per-class F1 ranging from
0.895 to 0.963, suggesting that task-type classification is reliable
enough.

Human evaluation confirms that system-level improvements translate to user 
experience. \sys\ obtains the highest score on every satisfaction dimension in 
Table~\ref{tab:satisfaction}, with GPT-5.5 as the strongest baseline at 4.2 
overall. AutoGen's low naturalness score of 3.2 reflects conversational disruption 
from tool-augmented orchestration. Its lower Accuracy and Completeness relative to 
Qwen3-32B reflect context truncation across inter-agent handoffs.

\subsection{Robustness Analysis}
\label{sec:robustness}

Table~\ref{tab:robustness} evaluates four controlled conditions. Under direct responses both systems perform near ceiling. The largest gap emerges under inconsistent responses, where the client self-contradicts across turns, with \sys\ leading GPT-5.5 by 28.4 percentage points on Cls at 70.5 versus 42.1 and reducing Rep from 16.3 to 5.5. This reflects the pending-slot register retaining prior elicitation attempts and resolving contradictions before belief-state update, whereas GPT-5.5 accepts the latest utterance as ground truth. Under deflective conditions the Rep gap is largest at 24.2 versus 11.4. Under underspecified conditions SFR$_1$ remains 93.8 versus 81.7, confirming that tier-priority schema enforcement recovers high-priority slots.

\begin{figure*}[t]
  \centering
  \includegraphics[width=0.9\textwidth]{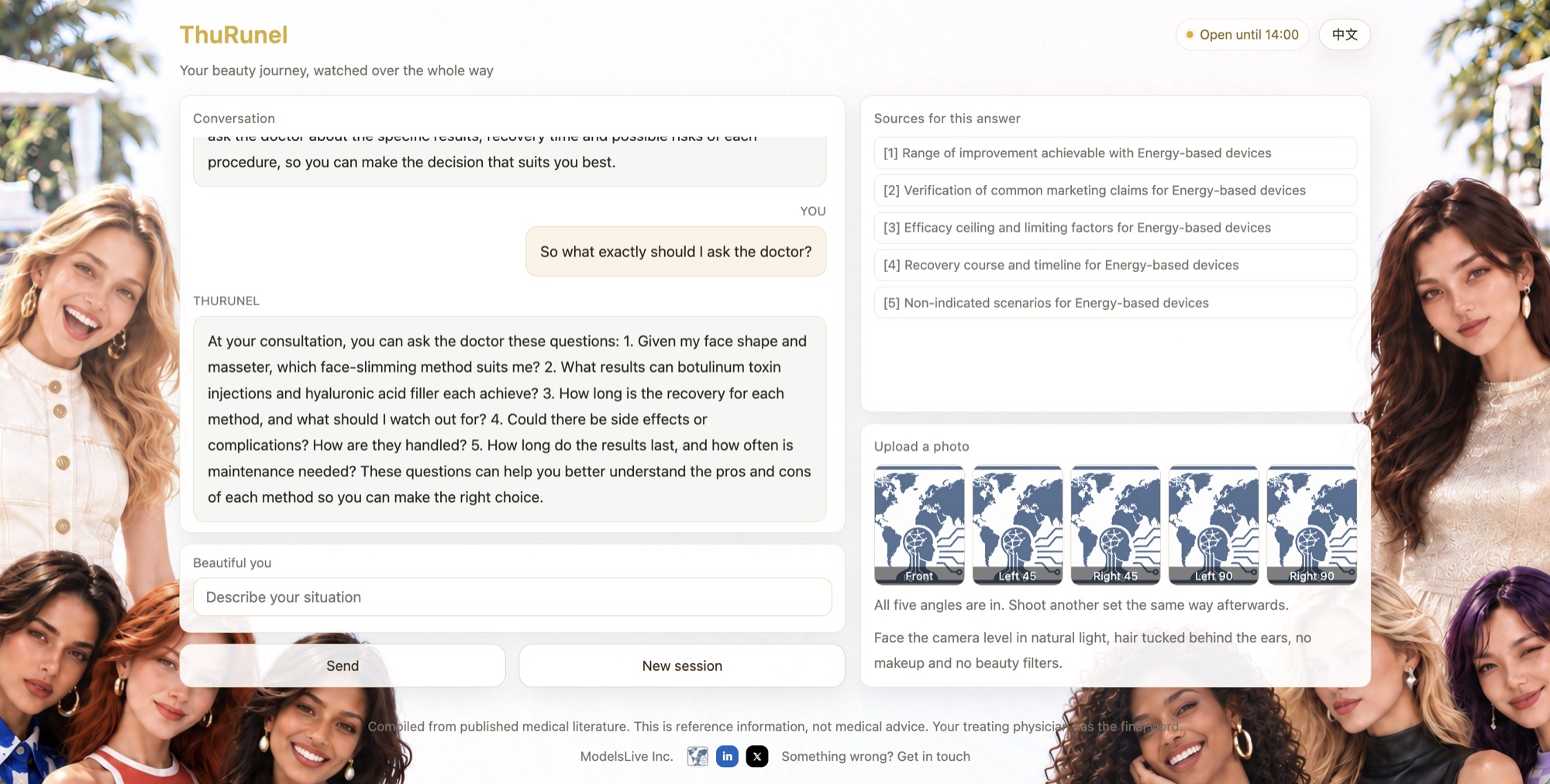}
  \caption{ThuRunel's public web interface. The conversation panel (left) lists questions for the client to bring to the consultation; the side panels (right) list the knowledge-base entries cited in this answer and the five-angle photo upload.}
  \label{fig:deploy}
\end{figure*}

\subsection{Case Study}
\label{sec:casestudy}

We show four pivotal turns from the session of Xiaohan, a 20-year-old client seeking non-surgical facial slimming. The full nine-turn dialogue is in Appendix~\ref{app:casefull}. We also compare the outputs of \sys\ and GPT-5.5. In Turn 1, \sys\ defers treatment content and begins profile elicitation while GPT-5.5 immediately lists treatment options without collecting any typed slot. In Turn 4, the client mentions sensitive skin as a casual aside alongside her budget; \sys\ recognizes this as a colloquial non-answer to the pending contraindication slot and issues a targeted follow-up, whereas GPT-5.5 treats it as a general preference and does not probe further. In Turn 5, the client confirms a prior hyaluronic acid reaction; \sys\ records it and flags it for specialist review, while GPT-5.5 acknowledges the reaction but immediately offers clinical guidance, crossing into specialist scope without structured recording. In Turn 9, \sys\ generates a compressed, structured brief implementing $\mathcal{T}_b$, with contraindication flag and specialist routing; GPT-5.5 produces a conversational summary that mixes advisory and clinical content without structured handoff or routing.

\section{Deployment}
\label{sec:deployment}

ThuRunel is publicly available as a bilingual Chinese and English web application, with no operator in the loop (Figure~\ref{fig:deploy}; further views in Appendix~\ref{app:deployed}). The evaluation above targets a clinic-facing setting, in which the agent's output is a brief routed to a specialist. The public deployment is an extended version of the pre-visit phase: the model and its adapters are unchanged, and the client-facing behavior is extended through the system prompt and an expanded knowledge base, so that the advisor serves prospective clients directly, before they have chosen a procedure or a clinic. No retraining was needed. Of the two phases evaluated above, the pre-visit phase is the one now publicly deployed; the in-clinic phase depends on integration with a clinic's consultation workflow and is not part of the public service. The deployment keeps the four decoupling decisions of Section~\ref{sec:problem} and changes who receives the handoff.

\paragraph{Decoupling from the client side.}
\emph{What to ask.} ThuRunel elicits the client's situation through a guided self-assessment, for example measuring a few key facial dimensions or pressing on an area to tell whether its fullness comes from bone or fat, and optionally through photos from five standard angles (front, left and right 45 degrees, left and right 90 degrees); a client who prefers not to upload can simply continue the conversation. \emph{What to resolve.} Within the dialogue it resolves advisory-scope questions: what a procedure is, what it can and cannot improve, where the upper bound of achievable improvement lies, and what the risks and recovery course are. \emph{What to forward.} In place of a brief routed to a specialist, the session closes with suitable directions for improvement and a list of questions the client should raise at the consultation, so that the specialist-scope decision stays with the doctor and the client arrives able to follow it. \emph{When to stop.} Diagnoses, treatment conclusions, pricing, and individual treatment plans are declined explicitly rather than improvised, and the agent recommends no clinic, doctor, or procedure.

\paragraph{Grounded knowledge with provenance.}
For this setting, the service catalog and case repository of Section~\ref{sec:data} are replaced by a curated knowledge base of 324 entries covering common procedures for the eyes, nose, facial contour, and skin, as well as body contouring, each maintained in both languages and annotated with its source. Entries are organized by the questions a client needs answered, such as the efficacy ceiling and limiting factors of a procedure, its recovery course and timeline, non-indicated scenarios, signs of overtreatment, warning signs of urgent complications, and verification of common marketing claims; for the last type, an entry states only the verified facts and does not reproduce the misleading claim itself. Every substantive answer lists, in a side panel, the entries actually cited in that turn.

\paragraph{Two access modes.}
The knowledge base is compiled into the web page and can be searched at any time without loading the model. The full conversation, which runs the dialogue framework and the fine-tuned adapters of Section~\ref{sec:system}, opens in daily windows, because the model shares a single GPU with other services. Separating the two modes keeps factual lookup always available while confining the cost of generation to the windows in which the full advisory flow is offered.

\paragraph{Serving.}
The Qwen2.5-32B-Instruct backbone is quantized to 4 bits and served with the adapters on one AWS g5.xlarge instance with a single A10G GPU of 24\,GB. Every conversation turn, uploaded photo, and set of cited sources is persisted, which also allows demonstrations and screenshots to be replayed from real sessions instead of being regenerated by the model.

\section{Conclusion}
\label{sec:conclusion}

We have presented the dynamic decoupling paradigm and its instantiation in \sys. The paradigm defines a role for an AI advisory agent that existing systems do not implement: an early-phase advisor that knows what it can resolve autonomously, knows when to stop eliciting, and produces a brief that enables the specialist to focus on consequential judgment. Experiments show that \sys\ outperforms
all baselines across both automatic and human evaluation metrics, and that teacher synthesis is an important contributor of dynamic decoupling
capability.
The public deployment shows that the same decisions carry over when the handoff goes to the client instead of the specialist. Future work will extend \sys\ to additional advisory domains by exchanging the slot schema and knowledge base.

\section*{Limitations}
\sys\ is evaluated in a single domain and language, Chinese medical aesthetics advisory dialogue, using simulated client sessions. The slot schema and knowledge bases are hand-specified for this domain, and the teacher synthesis protocol relies on a single backbone model, so results may vary with different base or teacher models. The full two-phase system is evaluated offline, while only the pre-visit phase is currently deployed; the in-clinic phase and the effect on real-world advisory outcomes will be measured as real usage accumulates.

\bibliographystyle{ACM-Reference-Format}
\bibliography{refs}

\appendix




\section{Case Study: Full Dialogue}
\label{app:casefull}

The following presents nine turns from the advisory session for Xiaohan, a 20-year-old client seeking non-surgical facial slimming, comparing \sys\ and GPT-5.5. Each figure corresponds to one turn in the deployed system. Client utterances are identical across both conditions. GPT-5.5 is given a domain system prompt describing the medical aesthetics context.

Figures~\ref{fig:t1}--\ref{fig:t3} cover the out-of-clinic intake phase. Figure~\ref{fig:t1} shows the opening turn, where \sys\ defers treatment content and begins profile elicitation while GPT-5.5 immediately lists treatment options. Figure~\ref{fig:t2} shows knowledge delivery: \sys\ answers the efficacy question and redirects to recovery-window elicitation while GPT-5.5 recommends treatment and moves to booking. Figure~\ref{fig:t3} shows continued elicitation: \sys\ follows up on recovery preference and asks about budget while GPT-5.5 again pushes toward booking. Figure~\ref{fig:t4} shows the contraindication turn: the client mentions sensitive skin as a casual aside; \sys\ issues a targeted pending-slot register follow-up while GPT-5.5 treats it as a general preference. Figure~\ref{fig:t5} shows the contraindication confirmed: \sys\ records the prior hyaluronic acid reaction and flags it for the specialist while GPT-5.5 notes it and moves to summarization. Figure~\ref{fig:t6} shows the booking turn: \sys\ executes a programmatic booking confirmation and forwards the intake profile while GPT-5.5 declines the request, stating that it cannot book appointments or operate a clinic. Figures~\ref{fig:t7}--\ref{fig:t9} cover the in-clinic phase. Figure~\ref{fig:t7} shows the photo upload turn: \sys\ generates photo-conditioned plan options grounded in the service catalog while GPT-5.5 produces a generic suggestion without access to the prior session. Figure~\ref{fig:t8} shows case retrieval: \sys\ surfaces comparable historical cases for expectation alignment while GPT-5.5 responds with safety advice about the prior reaction rather than expectation information. Figure~\ref{fig:t9} shows the specialist brief: \sys\ generates a compressed, structured brief with contraindication flag and specialist routing while GPT-5.5 produces a conversational summary mixing advisory and specialist content.


\begin{figure}[!htbp]
\begin{tcolorbox}[colback=lightgray, colframe=darkblue!50, fontupper=\small]
\textbf{Client:} Hello. I would like to learn about methods for making my face look slimmer.

\medskip
\textbf{\sys:} Hello! I am a medical aesthetics intake assistant. May I ask your name?

\medskip
\textbf{GPT-5.5:} There are several ways people make the face appear slimmer, ranging from temporary styling techniques to longer-term cosmetic procedures. What works best depends on your facial structure, body composition, age, and how dramatic a change you want.
\end{tcolorbox}
\caption{Turn 1: opening. \sys\ defers treatment content and begins structured profile elicitation. GPT-5.5 lists possible methods and asks about facial structure, body composition, and desired change, but does not collect name, age, or any typed slot.}
\label{fig:t1}
\end{figure}

\begin{figure}[!htbp]
\begin{tcolorbox}[colback=lightgray, colframe=darkblue!50, fontupper=\small]
\textbf{Client:} My name is Xiaohan. I am 20 years old. Which facial slimming method has better results?

\medskip
\textbf{\sys:} Results vary by individual. Surgical methods tend to produce more pronounced and lasting changes, while non-surgical options are gentler with shorter recovery. Do you have any preferences around recovery time?

\medskip
\textbf{GPT-5.5:} Hi Xiaohan. At age 20, your face still has natural youthful fullness, so the “best” method depends on what is actually creating the wider appearance. Different causes respond to different treatments.
\end{tcolorbox}
\caption{Turn 2: knowledge delivery. \sys\ answers the efficacy question and immediately redirects to recovery-window elicitation, implementing $\mathcal{T}_a$. GPT-5.5 discusses how the cause of facial width determines the best method, providing knowledge without advancing slot coverage.}
\label{fig:t2}
\end{figure}

\begin{figure}[!htbp]
\begin{tcolorbox}[colback=lightgray, colframe=darkblue!50, fontupper=\small]
\textbf{Client:} I would prefer a shorter recovery period.

\medskip
\textbf{\sys:} Understood. We can focus on non-surgical options such as jaw-slimming injections and radiofrequency tightening. Do you have a budget range in mind?

\medskip
\textbf{GPT-5.5:} If short recovery time is important, the best options are usually the non-surgical ones. Given your age and preference for minimal downtime, masseter Botox is often the first thing people consider if the jaw area is the main source of width. If your cheeks are naturally fuller, conservative approaches are usually safer at 20 because facial volume changes a lot over the next decade.
\end{tcolorbox}
\caption{Turn 3: continued elicitation. \sys\ confirms the recovery preference and elicits budget before advancing. GPT-5.5 recommends masseter Botox specifically, a treatment recommendation made before contraindications have been assessed.}
\label{fig:t3}
\end{figure}

\begin{figure}[!htbp]
\begin{tcolorbox}[colback=lightgray, colframe=darkblue!50, fontupper=\small]
\textbf{Client:} Budget around RMB 15,000. Skin-wise I am just sometimes a bit sensitive.

\medskip
\textbf{\sys:} You mentioned sensitive skin. Has your skin ever had a strong reaction to an injectable treatment, such as significant swelling or lasting redness afterward?

\medskip
\textbf{GPT-5.5:} With a budget of around RMB 15,000 and mild skin sensitivity, you have access to several good low-downtime options in China or similar markets. The safest and most effective choice still depends on what is making your face appear wider.
\end{tcolorbox}
\caption{Turn 4: contraindication elicitation. The client mentions sensitive skin alongside her budget. \sys\ recognizes this as a colloquial non-answer to the pending contraindication slot and issues a targeted follow-up. GPT-5.5 treats it as a general preference and notes that the best choice still depends on what is causing the wider appearance, neither resolving the slot nor probing further.}
\label{fig:t4}
\end{figure}

\begin{figure}[!htbp]
\begin{tcolorbox}[colback=lightgray, colframe=darkblue!50, fontupper=\small]
\textbf{Client:} Actually yes, I tried hyaluronic acid a few years ago and had a bad reaction.

\medskip
\textbf{\sys:} That is really important to know. A prior reaction to hyaluronic acid will shape which approaches are appropriate for you. I have noted this in your record and our specialist will want to discuss this carefully.

\medskip
\textbf{GPT-5.5:} That's important to know. A bad reaction to hyaluronic acid (HA) filler changes how you should approach facial treatments, especially injections.
\end{tcolorbox}
\caption{Turn 5: contraindication confirmed. \sys\ records the prior hyaluronic acid reaction and flags it for specialist review. GPT-5.5 acknowledges the reaction and immediately offers clinical guidance on how to approach treatments, crossing into the specialist scope without recording the contraindication in a structured format.}
\label{fig:t5}
\end{figure}

\begin{figure}[!htbp]
\begin{tcolorbox}[colback=lightgray, colframe=darkblue!50, fontupper=\small]
\textbf{Client:} Okay, please arrange a booking. Can I come in tomorrow morning at ten?

\medskip
\textbf{\sys:} Tomorrow at ten is confirmed. Your appointment has been booked and your intake profile has been forwarded to the in-clinic workspace. Please check in at the front desk on arrival.

\medskip
\textbf{GPT-5.5:} I appreciate that you want to move forward, but I need to clarify something important: I'm an AI assistant. I cannot book appointments, provide medical services, or operate a clinic.
\end{tcolorbox}
\caption{Turn 6: booking confirmation. \sys\ executes a programmatic booking confirmation and forwards the structured intake profile. GPT-5.5 declines the booking request, clarifying that it is an AI assistant that cannot book appointments or operate a clinic. This reflects a hard boundary in GPT-5.5 that prevents it from completing the advisory workflow regardless of context.}
\label{fig:t6}
\end{figure}


\begin{figure}[!htbp]
\begin{tcolorbox}[colback=lightgray, colframe=darkblue!50, fontupper=\small]
\textbf{Client:} [uploads facial photo]

\medskip
\textbf{\sys:} Based on your photo and the profile from your earlier session, we have three treatment options. The conservative plan uses a smaller dosage with recovery of approximately 3--5 days, estimated at RMB 2000--3000. The balanced plan uses a medium dosage with recovery of 5--7 days, estimated at RMB 3000--4000. The precision plan uses a larger dosage with recovery of 7--10 days, estimated at RMB 4000--5000.

\medskip
\textbf{GPT-5.5:} Xiaohan, based on this photo, your face looks soft oval/round with youthful cheek fullness, not obviously a very square “masseter-heavy” jaw. So for a slimmer look with short recovery, I would be conservative.
\end{tcolorbox}
\caption{Turn 7: photo upload and plan options. \sys\ inherits the out-of-clinic profile and generates photo-conditioned, dosage-specific plan options grounded in the service catalog $\mathcal{K}_S$. GPT-5.5 analyzes the photo independently and recommends a conservative approach based on facial shape, but has no access to the prior session and cannot incorporate the client profile, budget, or contraindication history into its response.}
\label{fig:t7}
\end{figure}

\begin{figure}[!htbp]
\begin{tcolorbox}[colback=lightgray, colframe=darkblue!50, fontupper=\small]
\textbf{Client:} What do results typically look like for this kind of treatment?

\medskip
\textbf{\sys:} The closest reference case in our repository involves a client with facial contour concerns who sought a slimming and lifting effect. Treatment applied was jaw-slimming injection combined with thread lifting. Please refer to the before-and-after images in the right panel.

\medskip
\textbf{GPT-5.5:} Because you had a bad reaction to hyaluronic acid before, tell the doctor clearly. Botox is a different product from HA filler, but your history still means you should choose a careful, licensed injector and start conservatively.
\end{tcolorbox}
\caption{Turn 8: case retrieval for expectation alignment. \sys\ retrieves the most similar historical case from $\mathcal{K}_C$ and presents before-and-after reference images. GPT-5.5 responds to the results question with safety advice about the prior hyaluronic acid reaction rather than expectation information, and has no access to comparable cases.}
\label{fig:t8}
\end{figure}

\begin{figure}[!htbp]
\begin{tcolorbox}[colback=lightgray, colframe=darkblue!50, fontupper=\small]
\textbf{Client:} I would like to proceed with the balanced plan.

\medskip
\textbf{\sys\ (specialist brief):} Client age: 20. Chief concern: facial slimming, non-surgical preference. Contraindication: prior hyaluronic acid reaction. Selected plan: balanced plan, medium dosage. Recovery window: 5--7 days. Estimated cost: RMB 3000--4000. Routing: Dr.~[matched specialist], facial aesthetics and injectable focus.

\medskip
\textbf{GPT-5.5:} Great, for your goals, budget, short recovery preference, and previous HA reaction, I’d suggest a balanced, conservative plan rather than aggressive reshaping.
\end{tcolorbox}
\caption{Turn 9: specialist brief. \sys\ generates a compressed brief implementing $\mathcal{T}_b$, with the contraindication flag, selected plan, and explicit specialist routing. GPT-5.5 suggests a balanced conservative plan but does not produce a structured handoff document, does not route to a specialist, and mixes advisory content with clinical guidance in a single conversational response.}
\label{fig:t9}
\end{figure}

\section{Clinic-Facing Workflow}
\label{app:deployment}

\sys\ has completed pilot deployment validation in a medical aesthetics clinic,
covering a closed-loop pipeline from out-of-clinic reception to in-clinic
continuation and specialist consultation. The goals are to reduce missed leads
and repetitive communication, improve conversion and follow-up efficiency, and
shorten specialist decision time. The turns of this workflow are those shown in
Appendix~\ref{app:casefull}; this section describes how they fit together.

The out-of-clinic phase operates primarily as client self-service with human
customer service as a fallback. The system greets the client, unifies service
messaging, and reduces communication barriers to build initial trust
(Figure~\ref{fig:t1}). A multi-turn elicitation then captures the client's
primary concern and desired improvements, budget range, recovery window and
risk preference, and relevant medical history and contraindications
(Figures~\ref{fig:t2}--\ref{fig:t5}). The system confirms the preferred visit
time, collects contact details, and records an appointment for follow-up
(Figure~\ref{fig:t6}). At the end of the phase it generates a structured record
summarizing the key information and the likely service direction
(Figure~\ref{fig:app_record}), which is pushed to the in-clinic consultant
workspace.

\begin{figure}[!htbp]
\begin{tcolorbox}[colback=white, colframe=accentorange!60, fontupper=\footnotesize\ttfamily]
chief\_complaint: make the face look smaller\\
goals: [make the face smaller through non-surgical means]\\
concerns: [shorter recovery]\\
interested\_projects: [jaw-slimming injection]\\
basic\_info: \{age: 20\}\\
history: \{allergy: none, medical\_history: none\}\\
appointment: \{date: 2026-01-07, time: 10:00, raw\_text: "10 a.m. tomorrow"\}\\
in\_clinic: false\\
summary: "The client wants a smaller face, prefers non-surgical methods and a short recovery, is interested in jaw-slimming injections, and has booked a consultation and examination for 10 a.m. tomorrow."
\end{tcolorbox}
\caption{Structured intake record generated at the end of the out-of-clinic phase and pushed to the in-clinic workspace, taken from a separate pilot session (fields translated from Chinese).}
\label{fig:app_record}
\end{figure}

The in-clinic phase also operates primarily as client self-service with human
consultants as fallback. The system ingests the out-of-clinic record and
booking information to avoid duplicate collection, and after the client uploads
a photo it drafts dosage-specific plan options for the consultant to explain
(Figure~\ref{fig:t7}). It retrieves similar historical cases to align effect
expectations, explain feasible scope, and surface relevant risk prompts
(Figure~\ref{fig:t8}). Finally, it generates a structured pre-specialist brief
and routes it, together with the booking information, to the matched
specialist (Figure~\ref{fig:t9}).

The specialist interface remains lightweight by design. The consultant completes
key information capture and initial explanation during the in-clinic phase and
\sys\ automatically generates a structured pre-specialist brief with booking
information, key constraints, and discussed directions. The specialist then
focuses on quickly reviewing the brief, confirming risks and preferences, and
finalizing the plan. Clients entering the specialist consultation have already
completed upfront information capture and boundary alignment, allowing the
specialist to focus time on consequential decisions and execution.

\section{Deployed System}
\label{app:deployed}

\begin{figure*}[!htbp]
  \centering
  \includegraphics[width=0.9\textwidth]{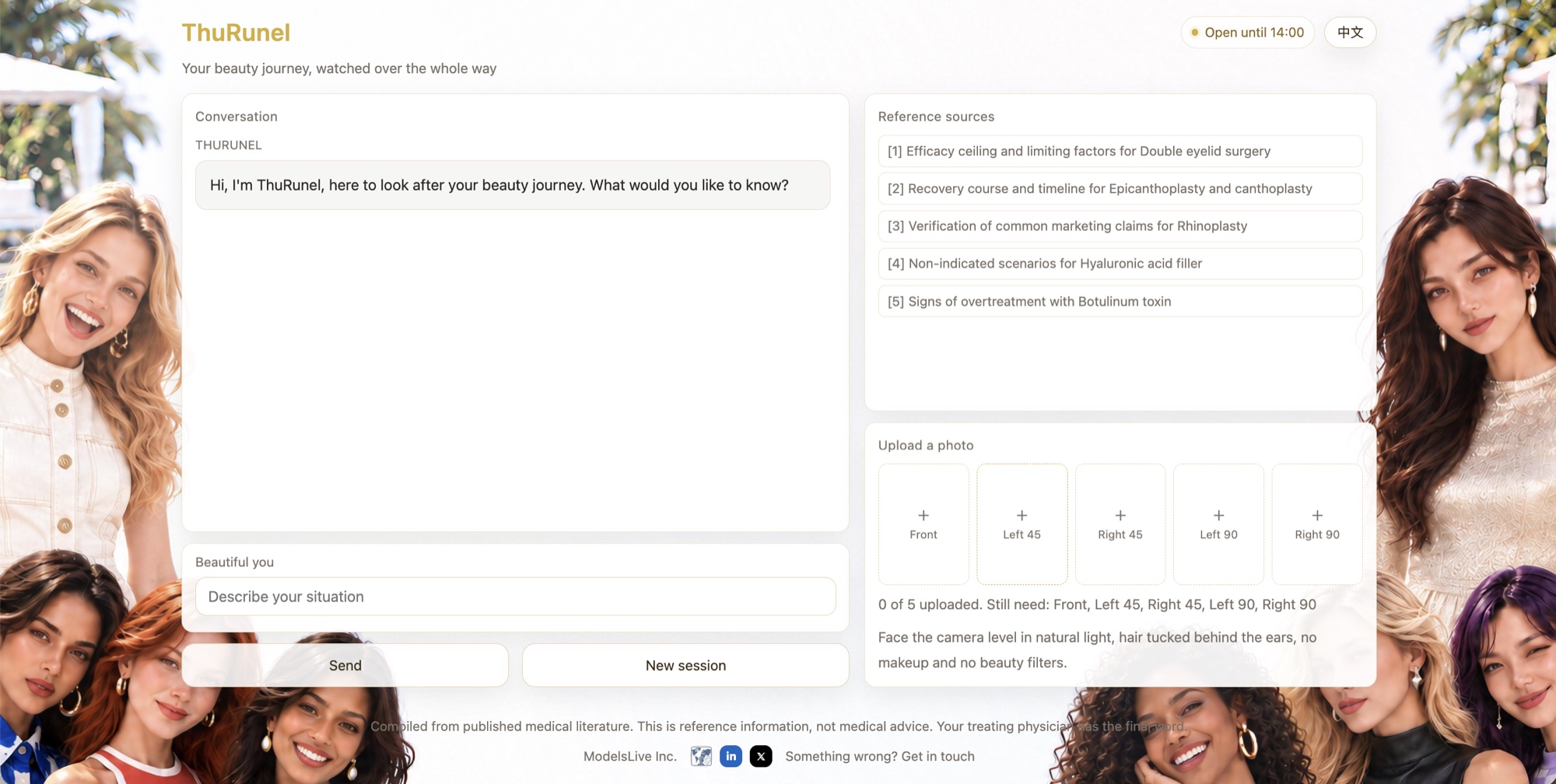}
  \caption{The ThuRunel web homepage, with the conversation panel, the reference-source panel, and the five-angle photo upload.}
  \label{fig:live1}
\end{figure*}

\begin{figure*}[!htbp]
  \centering
  \includegraphics[width=0.9\textwidth]{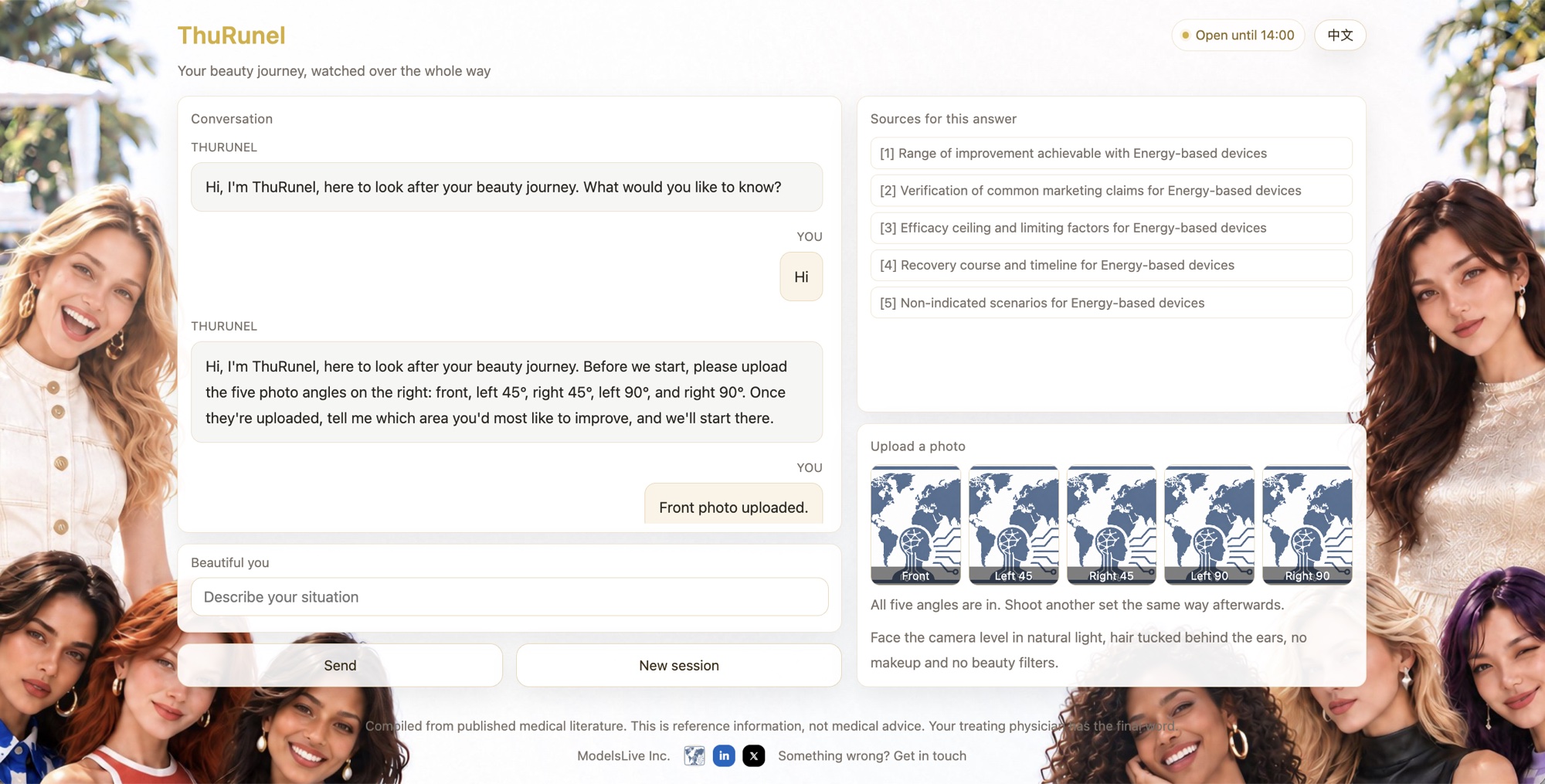}
  \caption{In the full conversation, ThuRunel guides the client to upload photos from five angles before discussing the area to improve.}
  \label{fig:live2}
\end{figure*}

\begin{figure*}[!htbp]
  \centering
  \includegraphics[width=0.9\textwidth]{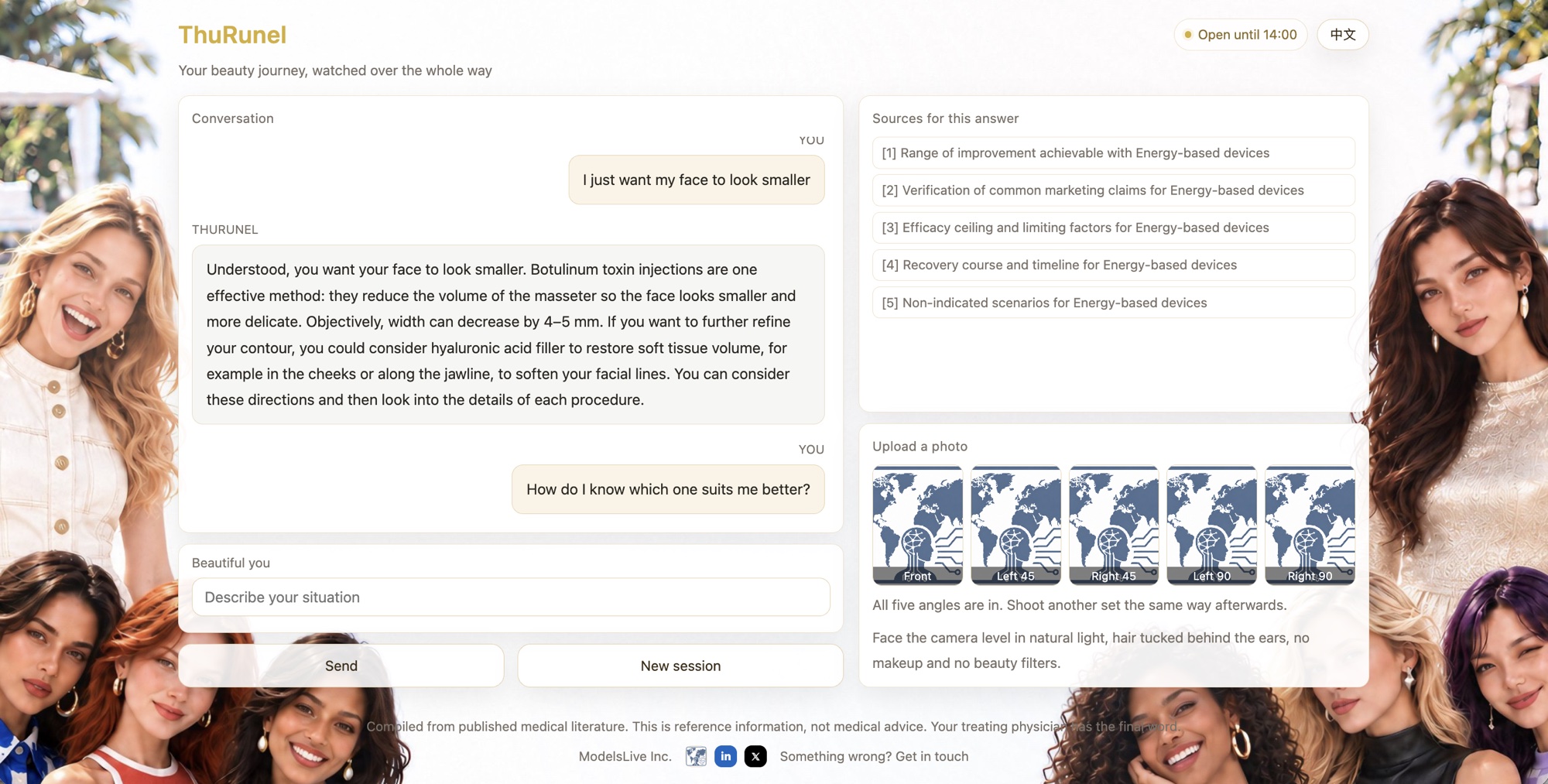}
  \caption{ThuRunel explains the upper bound the relevant procedures can reach, with the cited knowledge-base entries listed on the right.}
  \label{fig:live3}
\end{figure*}

\begin{figure*}[!htbp]
  \centering
  \includegraphics[width=0.9\textwidth]{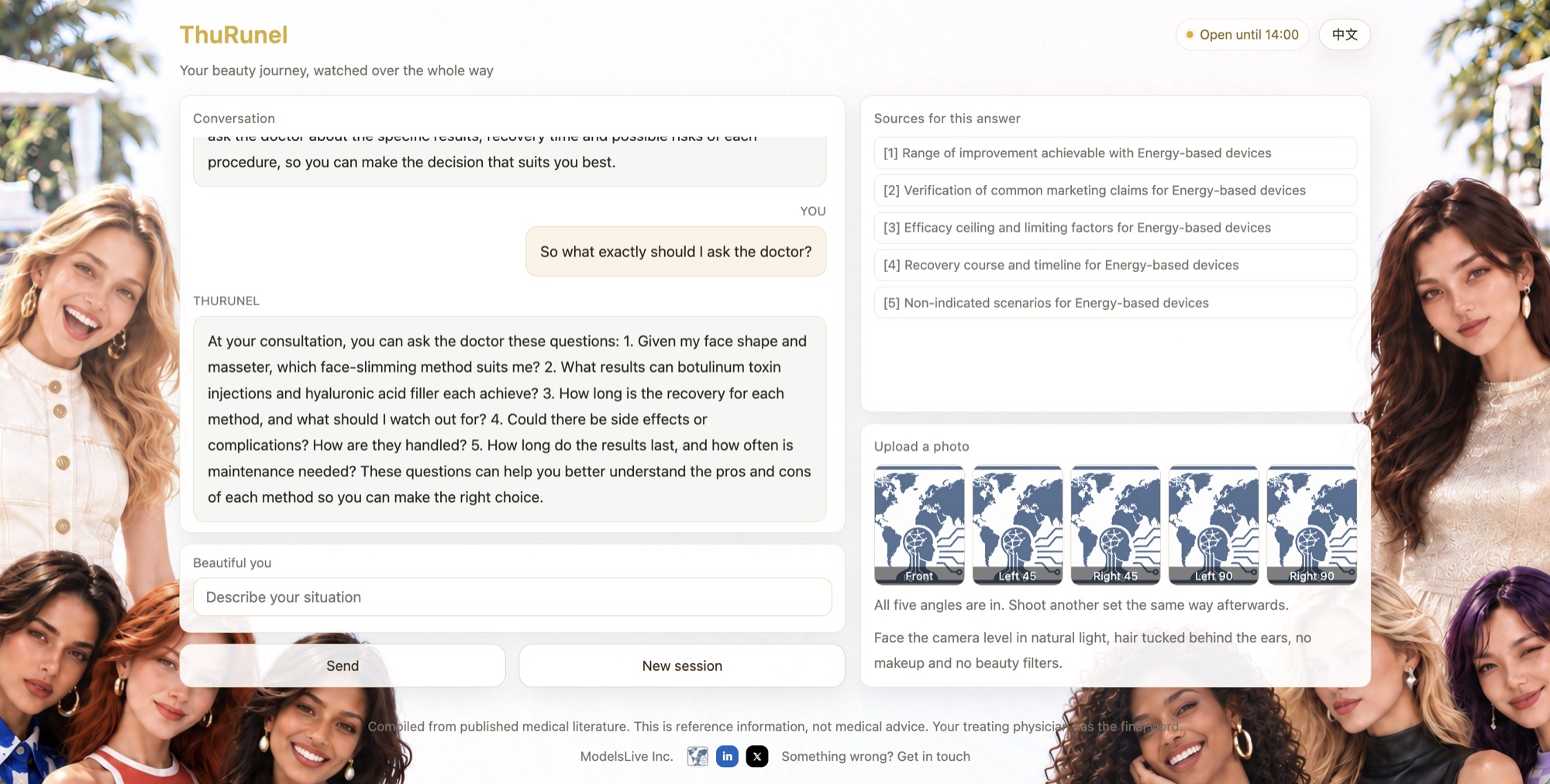}
  \caption{ThuRunel lists questions worth asking the doctor at the consultation, with the sources for this answer on the right.}
  \label{fig:live4}
\end{figure*}

\begin{figure*}[!htbp]
  \centering
  \includegraphics[width=0.9\textwidth]{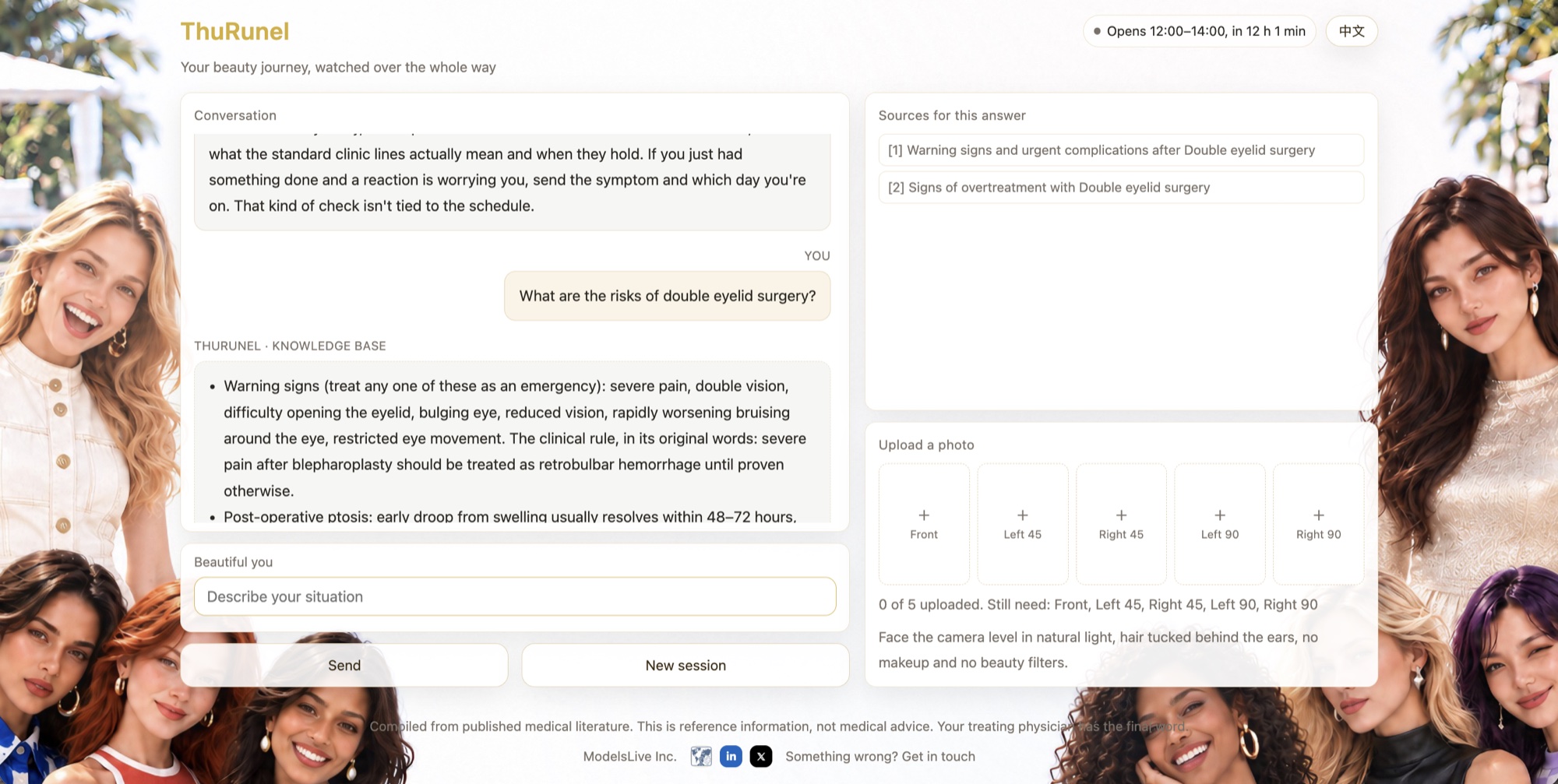}
  \caption{Knowledge-base mode: risks, recovery time, and warning signs can be looked up at any time, including outside the opening windows of the full conversation.}
  \label{fig:live5}
\end{figure*}

Figures~\ref{fig:live1}--\ref{fig:live5} show the publicly deployed ThuRunel described in Section~\ref{sec:deployment}, following one session from the homepage to the questions the client brings to the consultation, and the knowledge-base mode.

\end{document}